\pdfoutput=1

\documentclass[11pt]{article}

\usepackage{amsmath}
\usepackage{ACL2023}

\usepackage{comment}
\usepackage{times}
\usepackage{latexsym}

\usepackage{graphicx}
\newcommand{\ec}{$e$-$c$ }
\usepackage[T1]{fontenc}

\usepackage[utf8]{inputenc}

\usepackage{microtype}

\usepackage{inconsolata}

\definecolor{bulgarianrose}{rgb}{0.28, 0.02, 0.03}
\newcommand{\jo}[1]{\textcolor{bulgarianrose}{#1}}
\newcommand{\af}[1]{\textcolor{blue}{#1}}

\newcommand{\textlist}[1]{\textbf{#1}}

\title{
With a Little Push, NLI Models \textit{can} Robustly and Efficiently Predict Faithfulness
}

\author{Julius Steen
    \enspace
    Juri Opitz
    \enspace
    Anette Frank
    \enspace
    Katja Markert\\
  Department of Computational Linguistics \\
  Heidelberg University\\
  69120 Heidelberg, Germany \\
  {\tt (steen|opitz|frank|markert)@cl.uni-heidelberg.de}
}

\begin{document}
\maketitle
\begin{abstract}
Conditional language models still generate unfaithful output that is not supported by their input. These unfaithful generations jeopardize trust in real-world applications such as summarization or human-machine interaction, motivating a need for automatic faithfulness metrics. To implement such metrics, NLI models seem attractive, since they solve a strongly related task that comes with a wealth of prior research and data. But recent research 
suggests that NLI models require costly additional machinery to perform reliably across datasets, e.g., by running inference on a cartesian product of input and generated sentences, or supporting them with a question-generation/answering step.

In this work we show that pure NLI models \textit{can} outperform more complex metrics when combining task-adaptive data augmentation with robust inference procedures. We propose: (1) Augmenting NLI training data to
adapt  NL inferences to the specificities of faithfulness prediction in dialogue;
(2) Making use of both entailment and contradiction probabilities in NLI, and
(3) Using Monte-Carlo dropout during inference.
Applied to the TRUE benchmark, which combines faithfulness datasets across diverse domains and tasks, our approach strongly improves a vanilla NLI model and significantly outperforms previous work, while showing favourable computational cost. 
\end{abstract}

\section{Introduction}

Conditional language models suffer from a tendency to \textit{hallucinate} information \cite{maynez-etal-2020-faithfulness}, resulting in generations that are not faithful to their input documents, which limits the trustworthiness of such models. This raises a need for automatic faithfulness metrics. In this context, models trained on natural language inference (NLI) \citep{bowman-etal-2015-large} are attractive since, intuitively, a generation being \textit{faithful} implies it must be \textit{entailed} by the source \citep{falke-etal-2019-ranking}.

However, pure NLI models have seen mixed success in faithfulness evaluation \citep{falke-etal-2019-ranking, kryscinski-etal-2020-evaluating, wang-etal-2020-asking,maynez-etal-2020-faithfulness}. While in  recent evaluation on the TRUE benchmark \citep{honovich-etal-2022-true-evaluating}, which contains datasets from knowledge-grounded dialogue, summarization and paraphrasing, NLI-derived metrics perform best overall, they  require impractically large models, or costly additional machinery such as question generation and answering models at inference, while still showing robustness issues. Thus we ask: \textit{What is still needed for pure NLI models to perform robustly across faithfulness datasets -- while remaining cheap enough to serve as a lean and
practical evaluation tool?}

We enhance
a relatively small NLI model to make it work robustly across tasks in three ways:

\textlist{Task-Adaptive Data Augmentation.} 
 In NLI, a hypothesis must be fully entailed by its supporting premise.
 However, in faithfulness, not all parts of the generation always need to be grounded. 
 We identify an instance of this phenomenon in dialogue  where 
 parts of a turn can fulfill communicative functions such as hedging or establishing emotional connection
 and are often disregarded in faithfulness annotation. Hence, when applying NLI models to \textit{complete dialogue turns} that may include 
 statements irrelevant for grounding,
 we run a risk of
 producing incorrect unfaithfulness predictions.
 
To alleviate this issue, we propose a simple \textbf{data augmentation} method to adapt NLI models to genres 
where they need to be aware of statements that must be 
exempt from NLI-based faithfulness evaluation. Our approach is computationally attractive, 
as it avoids an increase of cost at inference time. 

\textlist{Integration of NLI Contradiction Scores.} 
Existing NLI faithfulness metrics typically use the entailment score for their
predictions \citep{honovich-etal-2022-true-evaluating, falke-etal-2019-ranking, kryscinski-etal-2020-evaluating}. However, \citet{menli} show that subtracting the contradiction score from the entailment score (referred to as \ec) can improve NLI performance
in certain evaluation tasks.
We show that there also is a strong positive effect of \ec for faithfulness prediction, and demonstrate that this is due to a high contradiction probability being a more
reliable predictor of unfaithfulness than low entailment probability.

\textlist{Monte-Carlo Dropout Inference.} Applying NLI models to faithfulness prediction
involves a domain shift from largely human-written data to automatically generated text.
To make NLI model scores more robust under this shift, we propose to use Monte-Carlo dropout during inference \citep{dropout}. This essentially creates a cheap \textit{ensemble} and has been shown to deal better with noisy labels \citep{dropout-noise}. This approach leads to consistent score improvements in our tasks.

The combination of all  modifications not only strongly improves over a baseline NLI model,
but also outperforms all other metrics on TRUE, on average, while being \textbf{cheaper} and \textbf{smaller}.\footnote{All code is available at \url{https://github.com/julmaxi/with_a_little_push}}

\section{Method Details}

\subsection{Task-adaptive Data Augmentation}

To illustrate that task requirements can be incompatible between faithfulness and NLI, consider the following instance from the Q2 dialogue corpus \citep{honovich-etal-2021-q2} that is labelled as faithful:
\begin{quote}
\textbf{Grounding:} American pancakes are similar to Scotch pancakes or drop scones.\\
\textbf{Generation:} yes , i love american pancakes , they are like scotch pancakes
\end{quote}
From an NLI perspective, the generation
is clearly not entailed, since the statement ``I love american pancakes'' is not supported by the input.

To better prepare an NLI system for such genre or task-specific cases, we 
manually curate a small list of statements that should not influence the faithfulness prediction. We augment NLI data from the ANLI corpus \citep{nie-etal-2020-adversarial} by adding a randomly chosen phrase from this set to each instance, while preserving the label. We then train an already fine-tuned NLI model on a concatenation of these augmented samples and original ANLI data. For training details see Appendix \ref{app:train}.

\subsection{Monte-Carlo Dropout}

To compute scores under Monte-Carlo dropout, we randomly sample $k$ dropout masks and compute the average of the model predictions. We set $k=15$, since preliminary experiments showed that performance did not profit from additional samples.

\begin{table*}

\centering\small\scalebox{0.98}{\begin{tabular}{|l|c|c|c||c|c|c||c|c|}
\hline
Method		&	Q2										&	SummacZS								&	T5 ANLI									&	\texttt{Base}									&	\texttt{-MC}							&  \texttt{All} 									&	\texttt{Eorig}								& \texttt{Eour} \\\hline

\multicolumn{9}{|l|}{\textbf{Summarization}} \\\hline

Frank		&	$_{\text{85.4}}$87.8$_{\text{90.0}}$	&	$_{\text{86.7}}$89.1$_{\text{91.1}}$	&	$_{\text{87.3}}$\textbf{89.4}$_{\text{91.2}}$	&	$_{\text{83.1}}$85.6$_{\text{88.0}}$	&	$_{\text{84.2}}$86.6$^{\dagger}_{\text{88.9}}$	&	$_{\text{85.5}}$87.7$^{\dagger}_{\text{89.8}}$	&	$_{\text{89.4}}$91.2$_{\text{93.0}}$	&					$_{\text{89.7}}$91.5$_{\text{93.2}}$			\\
MNBM		&	$_{\text{65.6}}$68.7$_{\text{71.7}}$	&	$_{\text{68.6}}$71.3$_{\text{74.1}}$	&	$_{\text{75.5}}$\textbf{77.9}$_{\text{80.2}}$	&	$_{\text{71.7}}$74.6$_{\text{77.4}}$	&	$_{\text{70.1}}$73.5$_{\text{76.6}}$	&	$_{\text{71.3}}$74.5$_{\text{77.4}}$	&	$_{\text{74.0}}$76.6$_{\text{79.4}}$	&									$_{\text{73.6}}$76.4$_{\text{79.2}}$			\\
SummEval	&	$_{\text{75.9}}$78.8$_{\text{81.4}}$	&	$_{\text{79.4}}$\textbf{81.7}$_{\text{83.9}}$	&	$_{\text{78.0}}$80.5$_{\text{83.0}}$	&	$_{\text{69.6}}$72.8$_{\text{75.8}}$	&	$_{\text{72.3}}$75.2$^{\dagger}_{\text{78.1}}$	&	$_{\text{73.2}}$76.1$^{\dagger}_{\text{78.8}}$	&	$_{\text{80.4}}$82.9$_{\text{85.4}}$	&					$_{\text{80.3}}$83.0$_{\text{85.3}}$			\\
QAGS-X		&	$_{\text{65.5}}$70.9$_{\text{76.2}}$	&	$_{\text{73.1}}$78.1$_{\text{82.9}}$	&	$_{\text{79.5}}$\textbf{83.8}$_{\text{88.2}}$	&	$_{\text{76.9}}$81.6$_{\text{86.5}}$	&	$_{\text{77.7}}$82.2$_{\text{86.8}}$	&	$_{\text{76.3}}$81.1$_{\text{85.4}}$	&	$_{\text{80.4}}$84.8$_{\text{88.9}}$	&									$_{\text{79.4}}$83.8$_{\text{88.0}}$			\\
QAGS-C		&	$_{\text{79.1}}$\textbf{83.5}$_{\text{87.9}}$	&	$_{\text{76.3}}$80.9$_{\text{85.2}}$	&	$_{\text{77.5}}$82.1$_{\text{86.7}}$	&	$_{\text{68.7}}$74.1$_{\text{79.3}}$	&	$_{\text{73.0}}$78.4$^{\dagger}_{\text{82.9}}$	&	$_{\text{73.2}}$78.0$^{\dagger}_{\text{82.9}}$	&	$_{\text{83.5}}$87.7$_{\text{91.3}}$	&					$_{\text{83.1}}$86.7$_{\text{90.3}}$			\\\hline

\multicolumn{9}{|l|}{\textbf{Dialogue}} \\\hline

BEGIN		&	$_{\text{77.2}}$79.7$_{\text{82.2}}$	&	$_{\text{79.2}}$82.0$_{\text{84.6}}$	&	$_{\text{80.3}}$\textbf{82.6}$_{\text{85.1}}$	&	$_{\text{77.5}}$80.4$_{\text{82.9}}$	&	$_{\text{75.7}}$78.5$_{\text{81.4}}$	&	$_{\text{76.4}}$79.3$_{\text{82.3}}$	&	$_{\text{84.1}}$86.2$_{\text{88.2}}$	&									$_{\text{82.1}}$84.7$_{\text{87.1}}$			\\
DialFact	&	$_{\text{85.4}}$86.1$_{\text{86.8}}$	&	$_{\text{83.3}}$84.1$_{\text{84.8}}$	&	$_{\text{76.8}}$77.7$_{\text{78.6}}$	&	$_{\text{81.0}}$81.8$^{*}_{\text{82.5}}$	&	$_{\text{91.3}}$91.8$^{*{\dagger}x}_{\text{92.3}}$	&	$_{\text{92.0}}$\textbf{92.5}$^{*{\dagger}x}_{\text{93.0}}$	&	$_{\text{89.9}}$90.4$_{\text{91.0}}$	&			$_{\text{94.1}}$94.5$^{x}_{\text{94.9}}$			\\
Q2			&	$_{\text{78.8}}$80.9$_{\text{83.0}}$	&	$_{\text{74.9}}$77.4$_{\text{79.7}}$	&	$_{\text{70.3}}$72.7$_{\text{75.2}}$	&	$_{\text{77.5}}$79.8$^{*}_{\text{82.0}}$	&	$_{\text{87.2}}$88.8$^{*{\dagger}x}_{\text{90.3}}$	&	$_{\text{87.8}}$\textbf{89.4}$^{*{\dagger}x}_{\text{90.9}}$	&	$_{\text{80.8}}$82.8$_{\text{84.9}}$	&			$_{\text{86.8}}$88.5$^{x}_{\text{90.1}}$			\\\hline

\multicolumn{9}{|l|}{\textbf{Paraphrasing}} \\\hline

PAWS		&	$_{\text{89.1}}$89.7$_{\text{90.3}}$	&	$_{\text{87.5}}$88.2$_{\text{88.7}}$	&	$_{\text{85.7}}$86.4$_{\text{87.1}}$	&	$_{\text{87.2}}$87.8$^{*}_{\text{88.4}}$	&	$_{\text{88.4}}$89.0$^{*\dagger}_{\text{89.6}}$	&	$_{\text{89.4}}$\textbf{90.0}$^{*\dagger}_{\text{90.5}}$	&	$_{\text{90.7}}$91.2$_{\text{91.7}}$	&			$_{\text{91.8}}$92.3$^{x}_{\text{92.8}}$			\\\hline\hline
\textbf{Avg}			&	$_{\text{79.7}}$80.7$_{\text{81.7}}$	&	$_{\text{80.4}}$81.4$_{\text{82.3}}$	&	$_{\text{80.6}}$81.5$_{\text{82.4}}$	&	$_{\text{78.8}}$79.8$_{\text{80.8}}$	&	$_{\text{81.7}}$82.7$^{\dagger}_{\text{83.6}}$	&	$_{\text{82.2}}$\textbf{83.2}$^{*\dagger}_{\text{84.1}}$	&	$_{\text{85.1}}$86.0$_{\text{86.8}}$	&	$_{\text{86.0}}$86.8$^{x}_{\text{87.7}}$ \\

\hline
\end{tabular}  }
\caption{
AUC scores for all models on TRUE. Small numbers indicate 95\% CIs computed via bootstrap. $*$ indicates statistically significant improvement over \texttt{T5}; $\dagger$:
statistically sign.\ improvement over \texttt{Base}; $^x$:  statistically sign.\ improvement over \texttt{Eorig} ($p \leq 0.05$, 
approximate randomization test). Best non-ensemble models in bold.
}

\label{tab:main_results}
\end{table*}

\section{Experimental Setup}

We run experiments on TRUE \citep{honovich-etal-2022-true-evaluating}, a benchmark that compiles 
a wide variety of faithfulness tasks in a standardized format. 
It contains 
summarization \citep{pagnoni-etal-2021-understanding, maynez-etal-2020-faithfulness, wang-etal-2020-asking, fabbri-etal-2021-summeval}, knowledge-grounded dialog \citep{honovich-etal-2021-q2, gupta-etal-2022-dialfact, begin}\footnote{TRUE uses an earlier variant of BEGIN that is described in \url{https://arxiv.org/pdf/2105.00071v1.pdf}} and paraphrasing \citep{zhang-etal-2019-paws} datasets.\footnote{
TRUE
also has a fact-checking part, which was not included in average metric performance. We also exclude it here, as our base NLI model was trained on parts of it.} Following recommendations in TRUE, we evaluate using Area under the ROC Curve (AUC).

As our \texttt{BASE} model, we use the DeBERTa-large \citep{deberta} model of \citet{laurer}, trained on MultiNLI \citep{williams-etal-2018-broad}, Fever-NLI \citep{thorne-etal-2018-fever}, ANLI \citep{nie-etal-2020-adversarial}, LingNLI \citep{parrish-etal-2021-putting-linguist} and WANLI \citep{wanli}. 
The metric \texttt{All} uses all three of our proposed modifications to \texttt{Base}. We also investigate a variant without MC dropout inference (\texttt{-MC}) as a more cost efficient alternative.

We compare to the 
strongest models on TRUE:

\textbf{\texttt{T5 ANLI} {\rm \citep{honovich-etal-2022-true-evaluating}}} is a T5-11B \citep{t5} model trained on ANLI.\footnote{The original \texttt{T5} model is also pretrained on GLUE \citep{wang-etal-2018-glue} and SuperGLUE \citep{superglue} data, which contains additional NLI data.}

\textbf{\texttt{SummacZS {\rm \citep{laban-etal-2022-summac}}}} evaluates an NLI model on all pairs of input and generated sentences and then averages maximum entailment probabilities for each generated sentence.

\textbf{\texttt{Q2 {\rm \citep{honovich-etal-2021-q2}}}} combines a question generation/answering pipeline with an NLI score.

Finally, \citet{honovich-etal-2022-true-evaluating} introduce a strong ensemble of these  3 methods (\texttt{Eorig}). To further verify our approach, we construct a new ensemble (\texttt{Eour}) by replacing \texttt{T5} with \texttt{All}.

\section{Results} \label{sec:results}

Table~\ref{tab:main_results} shows the AUC scores for each metric. Our model \texttt{All} not only significantly improves over \texttt{Base} on six out of nine corpora, but also significantly outperforms all other competitors on average, while being more computationally efficient. 

As expected, we find the biggest gains in dialogue, where the \texttt{All} model even outperforms \texttt{Eorig} on 2 out of 3 corpora. We do not improve on BEGIN, which is likely due to bias in the dataset construction, which we elaborate on in Section~\ref{sec:dialog}. On the summarization part, \texttt{All} improves significantly over \texttt{Base} on 3 out of 5 corpora, while not significantly harming performance on any corpus. However, it still falls short of the best models in TRUE. The strong showing of \texttt{T5} on these corpora suggests that this might be alleviated with a stronger base model.

Overall, a very similar behaviour is exhibited by \texttt{-MC}, presenting an attractive option when the added overhead of multiple samples is undesirable.

\texttt{Eour} is
on par with \texttt{Eorig}, despite
massively reduced costs; it
even significantly outperforms it on two dialog and the paraphrasing corpora.

We also investigate the performance of each individual modification to our model 
(Table~\ref{tab:ablation}). They all
improve average scores, while only leading to a notable decrease on BEGIN for both \ec and dialogue augmentations and on MNBM for \ec.

Outside of dialogue, we find that the augmentation methods have a positive impact on PAWS, as well as all summarization corpora that are at least partially based on summaries for the CNN/DM dataset \citep{hermann_teaching_2015} (Frank, QAGS-C, and SummEval).
While we do not have a definitive explanation for this phenomenon, we hypothesize that on these datasets our augmentations aid in making the model robust in the presence of noise or irrelevant context since our augmentations are label-neutral and must similarly be 'ignored' during training.

\begin{table}

\footnotesize\centering\begin{tabular}{|l|c|c|c|}
\hline

Corpus  &       +\ec    &       +MC       &  +Aug. \\\hline

Frank	&	$_{\text{-0.0}}$+0.3$_{\text{+0.5}}$	&	$_{\text{+0.1}}$+0.9$_{\text{+1.8}}$	&	$_{\text{+0.3}}$+1.0$_{\text{+1.7}}$\\
MNBM	&	$_{\text{-2.1}}$-0.8$_{\text{+0.5}}$	&	$_{\text{+1.4}}$+2.1$_{\text{+2.9}}$	&	$_{\text{-0.4}}$+0.0$_{\text{+0.6}}$\\
SummEval	&	$_{\text{+0.7}}$+1.0$_{\text{+1.3}}$	&	$_{\text{+0.1}}$+1.2$_{\text{+2.3}}$	&	$_{\text{+0.6}}$+1.6$_{\text{+2.6}}$\\
QAGS-X	&	$_{\text{-0.4}}$+0.3$_{\text{+0.9}}$	&	$_{\text{-1.5}}$-0.2$_{\text{+1.1}}$	&	$_{\text{-0.3}}$+0.9$_{\text{+2.1}}$\\
QAGS-C	&	$_{\text{+0.5}}$+1.2$_{\text{+2.0}}$	&	$_{\text{-1.6}}$-0.1$_{\text{+1.5}}$	&	$_{\text{+2.2}}$+3.5$_{\text{+5.0}}$\\\hline
BEGIN	&	$_{\text{-3.0}}$-1.1$_{\text{+0.6}}$	&	$_{\text{+0.0}}$+0.6$_{\text{+1.3}}$	&	$_{\text{-1.6}}$-1.0$_{\text{-0.5}}$\\
DialFact	&	$_{\text{+8.3}}$+9.1$_{\text{+9.9}}$	&	$_{\text{+1.1}}$+1.3$_{\text{+1.5}}$	&	$_{\text{+3.1}}$+3.3$_{\text{+3.5}}$\\
Q2	&	$_{\text{+5.1}}$+6.5$_{\text{+7.9}}$	&	$_{\text{-0.4}}$-0.0$_{\text{+0.4}}$	&	$_{\text{+3.5}}$+4.2$_{\text{+5.0}}$\\\hline
PAWS	&	$_{\text{+0.3}}$+0.4$_{\text{+0.5}}$	&	$_{\text{+1.1}}$+1.3$_{\text{+1.4}}$	&	$_{\text{+0.8}}$+0.9$_{\text{+1.0}}$\\\hline\hline
Avg	&	$_{\text{+1.6}}$+1.9$_{\text{+2.2}}$	&	$_{\text{+0.5}}$+0.8$_{\text{+1.1}}$	&	$_{\text{+1.4}}$+1.6$_{\text{+1.9}}$ \\\hline

\hline

\end{tabular}

\caption{
AUC differences for 
individual modifications of
\texttt{Base}. Small numbers:
95\% CIs 
(bootstrap resampling).
}
      \label{tab:ablation}

\end{table}        

\section{Analysis}

\subsection{Effect of Dialogue Adaptation} \label{sec:dialog}


We investigate whether the improvements via our augmentation approach are indeed due to them improving the handling of personal statements.

We use the occurrences of the pronoun \textit{I} in a generation as a proxy measure\footnote{We use spacy (\url{spacy.io}) for POS tagging to identify pronouns.} and compute its correlation with human labels and metrics (see Table~\ref{tab:i_corr}). On both Q2 and Dialfact, our proxy measure, while uncorrelated with human labels, is strongly correlated with the scores of both \texttt{Base} and \texttt{T5}. This indicates these metrics indeed tend to incorrectly reject generations with personal statements. \texttt{All} on the other hand reduces this dependency.

\begin{table}
    \centering
    \begin{tabular}{|l|c|c|c|}
    \hline
Method	&	(BEGIN)	&	Q2	&	DialFact\\\hline
\texttt{T5}	&	(-0.27)	&	-0.40	&	-0.13\\\hline
\texttt{Base}	&	(-0.28)	&	-0.32	&	-0.10\\
\texttt{All}	&	(-0.19)	&	-0.19	&	0.04\\\hline\hline
Gold Label	&	(-0.35)	&	-0.03	&	0.05\\
\hline
\end{tabular}
    \caption{Kendall's $\tau$ correlations of gold labels/system scores with first person pronoun occurrence. BEGIN shows a strong negative correlation which we attribute to model-induced dataset bias (see Appendix~\ref{app:sampling_bias_begin}).}
    \label{tab:i_corr}
\end{table}

Our results also help explain why \texttt{All} fails to improve on BEGIN, since BEGIN gold labels are negatively correlated with first person pronouns. This is likely due to a bias in dataset construction: The BEGIN dataset used in TRUE has generations from two models, one of which is both more likely to generate pronouns and more likely to generate unfaithful output (see Appendix~\ref{app:sampling_bias_begin}).

\subsection{Effect of integrating contradiction scores}

\begin{figure}
    \centering
    \includegraphics[width=1.0\linewidth]{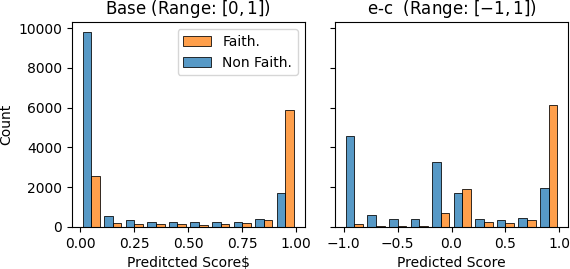}
    \caption{Histogram of the score distributions with and without \ec  for faithful and non-faithful instances.}
    \label{fig:true_dist}
\end{figure}

To isolate the effect of \ec we compare score distributions of \texttt{Base} and \texttt{Base+}\ec in Figure~\ref{fig:true_dist}.
The left-hand side of the figure shows that  in \texttt{Base} ca.\ 2700 faithful instances are predicted as non-entailed 
(i.e., $e$-score near 0), which implies they are 
labelled as contradictory or neutral.
\ec, on the other hand, further differentiates these instances into instances with high contradiction (negative \ec score) and high neutral probability (\ec score near 0). We 
observe that almost all 
low-scoring faithful generations are classified as neutral, whereas nearly all instances that are classified as contradictory are indeed unfaithful. Where \texttt{Base} has no way to make use of this information, \ec allows to reliably label contradictory instances as unfaithful.


\subsection{Cost comparison to other approaches}

\begin{table}
    \centering
    \scalebox{0.82}{\begin{tabular}{|l|c|c|c|c|}
    \hline
       Method  & AUC$\uparrow$ &Param$\cdot$$10^6$$\downarrow$ &  Model calls$\downarrow$ \\
       \hline
    SummacZS & 80.7& 355 & \#snt$\times$\#snt \\
    T5 ANLI & 81.5 & 11,000 & 1 \\
    Q2 & 81.4& 220 + 355 + 355 &  $\text{\#Q}\times(\text{Ql} + 2)$ \\
    \hline
    \texttt{-MC} & 82.7 & 350 & 1\\
    \texttt{All} & 83.2& 350 & 15 \\
    \hline
    \end{tabular}}
    \caption{Performance vs.\ cost analysis}
    \label{tab:cost_analysis}
\end{table}

There is increasing awareness of the resource-hun\-gri\-ness of deep learning \cite{strubell-etal-2019-energy}. Especially
for faithfulness,
cheap and reliable metrics are critical, given
rising demands for NLG in research and industry. 
Table \ref{tab:cost_analysis} shows
that our model
requires fewer parameters than any other metric, including a more than 30x reduction compared to T5. During inference our model always requires a constant number of calls which can be reduced to a single call when ablating MC dropout. On the other hand, the number of calls in SummacZS scales with the number of input and output sentences. Q2 needs  
to generate questions 
by calling an auto-regressive QG model $n$ times, where $n$ factors in the amount and length of questions (\#Q$\times$Ql), answer \#Q questions with the QA model and finally check \#Q answers with an NLI model ($\text{\#Q}\times 2$).

In sum, our model compares favourably with other approaches, while also allowing for a performance/cost tradeoff by forgoing MC dropout.

\subsection{Phrase Selection Robustness}

\begin{table}
\centering
\small\begin{tabular}{|l|c|c|c|c||c|}
\hline
& \multicolumn{4}{c||}{w/ Five Augmentations} & No Aug. \\
Dataset & Avg. & Std. & Min & Max & Avg.  \\\hline
Frank & 86.7$_{\text{-1.0}}$ & 0.4 & 85.8 &87.6 & 86.2  \\
MBNM & 74.4$_{\text{-0.1}}$ & 0.4 & 73.7 & 74.9 & 75.1\\
SummEval & 75.2$_{\text{-0.9}}$ & 0.5 & 74.5 & 76.0 & 74.3 \\
QAGS-X & 81.6$_{\text{+0.5}}$ & 0.5 & 80.8 & 82.4 & 80.7 \\
QAGS-C & 76.4$_{\text{-1.6}}$ & 0.8 & 74.7 & 77.9 & 75.2 \\\hline
DialFact & 92.1$_{\text{-0.4}}$ & 0.2 & 91.5 & 92.3 & 91.2 \\
BEGIN & 79.6$_{\text{+0.3}}$ & 0.5 & 79.0 & 80.6 & 80.9 \\
Q2 & 88.8$_{\text{-0.6}}$ & 0.3 & 88.1 & 89.2 & 86.3 \\\hline
PAWS & 89.7$_{\text{-0.3}}$ & 0.1 & 89.5 & 90.0 & 89.3 \\\hline\hline
\textbf{Avg.} & 82.7$_{\text{-0.5}}$ & 0.2 & 82.3 & 82.9 & 82.1 \\
\hline
\end{tabular}
\caption{
Results of our phrase selection robustness analysis. For each run, we sample five phrases, recreated our dataset and retrain our model. We repeat this process ten times and report the average, as well as the standard deviation, minimum and maximum scores of the runs. Small numbers indicate difference to the original scores. All results were computed using \ec and MC dropout. For better comparison, we also report the scores of a model without any augmentation (i.e. without any additional training) with \ec and MC dropout.
}
\label{tab:robustness}
\end{table}

To ensure that our augmentation is robust and not overly reliant on any particular choice of phrases, we repeat our dataset augmentation process multiple times with five randomly chosen augmentation phrases out of the original ten. We sample ten such datasets and retrain our model for each. Table \ref{tab:robustness} shows the average score, minimum and maximum score, as well as the standard deviation of the scores. We also report results of a model with both MC dropout and \ec but without any additional training and augmentations to directly quantify whether the augmentations are still helpful in their reduced form. This corresponds to applying MC dropout and \ec to \texttt{Base}.

As expected, we find that reducing
the variety of available phrases leads to a drop in performance across almost all datasets, compared to \texttt{All}. The only exception 
is BEGIN, where we instead see a slight improvement. This is likely to be related to the construction of BEGIN 
(see the discussion in Section \ref{sec:dialog}).

When comparing our limited augmentation models 
to the non-augmented model, we find that they
still outperform the non-augmented model
in almost all cases. In particular for Q2 and DialFact, for which
we expect the strongest impact of our augmentations,
we find that even the worst run still outperforms non-augmented model. This suggests that our augmentations can robustly adapt the model to the dialogue task.

Finally, we observe a relatively large drop in scores for all datasets that are at (least partially) derived
from CNN/DM (Frank, SummEval and QAGS-C). This 
mirrors 
our earlier observation in Section~\ref{sec:results} that these datasets profit from our augmentation procedure.

\section{Related Work}

Previous work on the utility of NLI for faithfulness led to mixed conclusions.
In summarization, \citet{falke-etal-2019-ranking} and \citet{kryscinski-etal-2020-evaluating} find out-of-the-box models have only limited utility in a faithfulness setting. In \citet{wang-etal-2020-asking}, an NLI model is outperformed by a question generation/answering (QA/QG)-based method. In contrast, \citet{maynez-etal-2020-faithfulness} find that a similar NLI model vastly outperforms a QA/QG metric on their data. In knowledge-grounded dialogue, \citet{begin}, \citet{gupta-etal-2022-dialfact} and \citet{honovich-etal-2021-q2} 
find out-of-the-box models 
underperform.

To improve NLI models for faithfulness in sum\-ma\-ri\-za\-tion, \citet{kryscinski-etal-2020-evaluating} propose FactCC, which is trained on artificially noised summaries. \citet{utama-etal-2022-falsesum} propose 
a controllable generation model to generate artificial faithfulness data.
In knowledge-grounded dialogue, \citet{begin} and \citet{gupta-etal-2022-dialfact} combine noising techniques to generate additional training data for NLI-based faithfulness models. In contrast to our work, these approaches a) generate training data from external sources, instead of directly augmenting NLI data, and b) do not explicitly focus on reconciling differences between NLI and faithfulness with their augmentation. Outside of augmentation-based approaches, \citet{goyal-durrett-2020-evaluating} propose to train NLI models to label faithfulness at the dependency arc level.



\section{Conclusion}

We have demonstrated that with a small number of focused adaptations, even a relatively small NLI model can robustly predict faithfulness. 
We have:
\begin{enumerate}
    \item Shown that NLI-based metrics can be incompatible with task-specific requirements and identified and fixed one such incompatibility in dialogue with an augmentation strategy.
    \item Demonstrated the importance of contradiction probability for scoring and that the underlying mechanism is the high reliability of NLI contradiction scores for detecting unfaithfulness
    \item Shown that using Monte-Carlo dropout improves metric performance.
\end{enumerate}

\noindent Our improved NLI model 
significantly improves over its baseline across many corpora and outperforms all competitors in average score on TRUE, while being much more efficient at inference.

Our work suggests that strong
improvements are possible for
NLI-based
faithfulness metrics, by combining
data augmentation with adapted
NLI score computation. We hope
this finding
will spurn 
advances
in cheap and robust NLI for faithfulness.


\section{Limitations}

Some of the  summarization datasets annotated for faithfulness are relatively small, which makes score estimates uncertain. Furthermore, many datasets contain only output from a limited number of generation systems, which makes it hard to properly account for potential biases towards certain generation systems that may confound scores (see \citet{pagnoni-etal-2021-understanding}). These concerns are, however, alleviated to some extent since we study trends across many independently created datasets, which makes it less likely for a single bias to persist in all of them. Furthermore the availability of generation and thus annotated faithfulness data limits our experiments to English. Finally, it remains unclear whether our results would still provide advantages when applied to larger models such as T5-11B, whose parameter count makes experimentation infeasible on the hardware available to us.

\section{Ethics Statement}

Faithfulness metrics help reduce the amount of incorrect information generated by NLG systems, reducing the risk associated which such generations. However, faulty or unreliable faithfulness metrics might cause harm by incorrectly classifying faithful content as unfaithful and vice versa.

We run all experiments on publicly available data that has been specifically constructed for faithfulness evaluation. The underlying publication has been published at a conference whose review process involved an ethics review. For a specific discussion of the human effort involved in creation of the datasets we refer the reader to the original publications.










\bibliography{anthology,custom}
\bibliographystyle{acl_natbib}

\appendix

\section{Augmentation Training Details} \label{app:train}

\subsection{Augmentation Phrases}

\begin{table}
    \centering
    \begin{tabular}{|c|}
   \hline \textbf{Introductory Statements} \\\hline
Here is what I know:\\    \hline
yep. Also \\    \hline
Sure! Here is what I know: \\    \hline
   \hline \textbf{Hedging} \\\hline
I am not sure, but \\    \hline
I am not sure but I do know that \\    \hline
I do not have information on this but \\    \hline
I think \\    \hline
I believe \\    \hline
   \hline \textbf{Sentiment} \\\hline
I love that! \\    \hline
I like that! \\    \hline
    \end{tabular}
    \caption{Manually curated list of dialogue phrases}
    \label{tab:phrases}
\end{table}


Table~\ref{tab:phrases} lists our manually curated list of phrases inserted during data augmentation. All phrases were derived via a small manual error analysis on the \texttt{Base} model.

We broadly divide our phrases into three categories: introductory statements, hedging, and sentiment statements. For each instance in ANLI, one random phrase from the list is prepended to the hypothesis. We use all three rounds of ANLI annotations. This results in 162,865 augmented instances which, together with the original ANLI instances, leads to a total of 325,730 training instances.

\subsection{Hyperparameters}

\begin{table}
    \centering
    \begin{tabular}{|l|c|}
    \hline
        Parameter & Val. \\\hline
        Warmup Ratio & 0.06 \\
        Weight Decay & 0.01 \\
        Effective Batch Size & 64 \\\hline
    \end{tabular}
    \caption{Hyperparameters}
    \label{tab:hparams}
\end{table}

Table~\ref{tab:hparams} lists the hyperparameter settings for our model. We use the same optimizer hyperparameters as \citet{laurer} except for an increased batch size and the learning rate. For the latter we tested three learning rates ($5e-6$, $5e-2$, $5e-1$) and select the one that provided the best loss on the augmented ANLI validation set. We initially ran models for 10,000 steps with a checkpoint every 1,000 steps and selected the checkpoint with the lowest loss on the augmented ANLI validation set. Later we reduced the number of training steps to 2,000 since we found we would usually select an early checkpoint as validation loss increased later in training, likely related to overfitting on the augmented data.

\subsection{Training}

We use the DeBERTa implementation in the huggingface transformers library \citep{wolf-etal-2020-transformers} and trained our model on a single node using two RX6800 GPUs, with one training run taking about three hours. Later experiments with fewer steps cut that time by 80\%.

\section{Dataset Bias in BEGIN} 
\label{app:sampling_bias_begin}

BEGIN is the only dialogue corpus on which first person pronoun occurrence shows a strong (negative) correlation with faithfulness (see Table~\ref{tab:i_corr}). 
Since there is nothing in the annotation guidelines that would explain this correlation, we instead hypothesize that this is the consequence of a model induced bias in the data.
Specifically, we hypothesize that one of the two models in BEGIN is (1) \textit{more} likely to generate personal statements and (2) \textit{less} likely to generate faithful responses.

To avoid confusion in the remainder of this section, we highlight that there are two variants of BEGIN: 
\begin{description}
     \item[BEGIN-v1] is the variant used in TRUE. It contains labeled generations by a fine-tuned GPT-2 base \citep{radford2019language} and a fine-tuned T5 base model \citep{t5} on the Wizard of Wikipedia dataset \citep{dinan2018wizard}.\footnote{The relevant data can be found at \url{https://raw.githubusercontent.com/google/BEGIN-dataset/5fa0cb0dde0e653d2016724a52a5ca27fe8b6a3f/dev_05_24_21.tsv}}
    \item[BEGIN-v2] is a more recent variant of BEGIN that is not part of TRUE. In addition to \textit{new} instances generated by T5 and GPT-2 it contains outputs from two additional models. It also has a revised annotation procedure. When we refer to BEGIN-v2, we exclusively mean the Wizard of Wikipedia subset.
\end{description}

Unfortunately, BEGIN-v1 does not allow us to retrieve which model generated which instance. This makes it impossible to directly investigate for model bias.
However, BEGIN-v2 includes outputs by the same two models, fine-tuned on the same data. Since we only need corpus level statistics to verify our assumptions, we conduct our analysis on the GPT-2 and T5 instances in BEGIN-v2.

To verify (1), we compute the correlation between a binary variable indicating which model generated each instance (T5: 0, GPT-2: 1) and first-person pronoun occurrence. We find a positive correlation (Kendall's $\tau$ wrt. to \textit{I}-pronoun occurrence: $0.18$, $p < 0.001$), indicating that GPT-2 generates outputs including more first-person pronouns.

To investigate whether GPT-2 is also more likely to be unfaithful, i.e. to verify (2), we compute the correlation between the binary model indicator variable and a faithfulness variable that is 1 when the output is labelled as \textit{Fully attributable} and 0 otherwise. We find a negative correlation (Kendall's $\tau$ wrt. to Faithfulness: $-0.25$, $p < 0.001$), supporting our hypothesis that GPT-2 is also overall less faithful. To ensure that this is not an effect of additional personal statements leading to more unfaithful generations, we conduct the same analysis only on instances where we identify no first-person pronouns. We find a similarly strong negative correlation of $-0.29$ ($p < 0.001$).

Our analysis shows that GPT-2 produces both overall less faithful outputs and more first-person pronouns than T5. Since BEGIN-v1 contains only outputs from T5 and GPT-2 this suggests that the root cause for the negative correlation between faithfulness label and first-person pronoun occurrence in BEGIN-v1 is model bias confounding faithfulness and first-person pronoun occurrence.

\subsection{Dataset Bias in BEGIN-v2}

We conduct a preliminary study to investigate whether similar biases also exist in BEGIN-v2. 

We observe that while BEGIN-v2 uses data from four dialogue systems, a majority of faithful generations is produced by a single system called \textsc{ctrl-Dialog} \citep{rashkin-etal-2021-increasing}. \textsc{ctrl-Dialog} is specifically trained to generate less subjective text, which we hypothesize might result in fewer first person pronouns. Since \textsc{ctrl-Dialog} also produces more faithful texts, this would lead to a negative correlation between faithfulness and first person pronouns, similar to what we observe on BEGIN-v1.

We verify this assumption by computing the correlation of a binary variable indicating an instance has been generated by \textsc{ctrl-Dialog} with a) the faithfulness labels on BEGIN-v2 and b) first-person pronoun occurrence. We find that an instance being generated by \textsc{ctrl-Dialog} is positively correlated with it having a \textit{faithful} label (Kendall $\tau$ w.r.t.\ faithfulness: 0.48, p$<0.001$) while being negatively correlated with the number of pronouns (Kendall $\tau$ w.r.t.\ \textit{I}-pronoun occurrence: -0.34, p$<0.001$). This suggests future evaluations on the BEGIN-v2 might run into similar bias issues.



\begin{table}[t]
    \centering
    \small\begin{tabular}{|l|r|r|r|}
    \hline
Corpus & Faith. & Non. Faith & Total \\\hline
Frank	&	223 (33.2\%)	&	448 (66.8\%)	&	671\\
MNBM	&	255 (10.2\%)	&	2245 (89.8\%)	&	2500\\
SummEval	&	1306 (81.6\%)	&	294 (18.4\%)	&	1600\\
QAGS-X	&	116 (48.5\%)	&	123 (51.5\%)	&	239\\
QAGS-C	&	113 (48.1\%)	&	122 (51.9\%)	&	235\\\hline
BEGIN	&	282 (33.7\%)	&	554 (66.3\%)	&	836\\
DialFact	&	3341 (38.5\%)	&	5348 (61.5\%)	&	8689\\
Q2	&	628 (57.7\%)	&	460 (42.3\%)	&	1088\\\hline
PAWS	&	3539 (44.2\%)	&	4461 (55.8\%)	&	8000\\\hline
\end{tabular}
    \caption{Dataset statistics for all constituent corpora in TRUE}
    \label{tab:true_stats}
\end{table}

\section{Dataset Statistics}

We report the number of instances, as well as the class distribution of TRUE in Table~\ref{tab:true_stats}.

\end{document}